# A Survey on Deep Learning-based Gaze Direction Regression: Searching for the State-of-the-art

Franko Šikić, Donik Vršnak, Sven Lončarić
University of Zagreb Faculty of Electrical Engineering and Computing, Zagreb, Croatia
Email: franko.sikic@fer.unizg.hr

## ABSTRACT

In this paper, we present a survey of deep learning-based methods for the regression of gaze direction vector from head and eye images. We describe in detail numerous published methods with a focus on the input data, architecture of the model, and loss function used to supervise the model. Additionally, we present a list of datasets that can be used to train and evaluate gaze direction regression methods. Furthermore, we noticed that the results reported in the literature are often not comparable one to another due to differences in the validation or even test subsets used. To address this problem, we re-evaluated several methods on the commonly used in-the-wild Gaze360 dataset using the same validation setup. The experimental results show that the latest methods, although claiming state-of-the-art results, significantly underperform compared with some older methods. Finally, we show that the temporal models outperform the static models under static test conditions.

## 1. INTRODUCTION

Gaze estimation is a research field with a centuries-old history. As early as 1849, Emile du Bois-Raymond discovered a relationship between electric potential and eyeball movements [1]. Modern-day gaze estimation methods can generally be divided into model-based and appearance-based methods. Model-based methods use a geometric eye model, whereas appearance-based methods directly use eye or head images to estimate the gaze direction or target location [2]. In recent years, scientific research has been much more focused on appearance-based methods as they do not require a calibration process and can work with images of lower resolution. In general, appearance-based methods can be divided into point-of-gaze (PoG) estimation and gaze direction regression methods. The former attempts to estimate the point in the image frame at which the human looks, whereas the latter attempts to estimate a vector that shows the direction of the human gaze from the eye or head toward the target. PoG methods estimate the gaze point only for in-frame points, whereas vector regression methods can estimate gaze even if the PoG is outside the frame, thus making them more robust and appropriate for in-the-wild images.

In this paper, we focus solely on deep learning-based methods for gaze direction regression. Throughout our research, we noticed that the results that have been reported in various papers cannot be compared one to another because the validation process is somehow changed or performed on a different data subset, thus making the claimed state-of-the-art results invalid. Therefore, we decided to re-evaluate numerous published methods on a commonly used dataset using the same validation protocol to find an actual state-of-the-art method. Furthermore, we analyze how static test conditions affect the results of temporal gaze regression models to further understand the generalization capabilities of such models. Overall, to the best of our knowledge, our contribution is three-fold: (a) we are the first to present a survey that focuses solely on deep learning-based gaze direction regression methods, (b) we are the first to address the problem of invalid comparisons between the reported results of published methods, and (c) we are the first to show that temporal models achieve better results than static models under static test conditions.

Following the introduction, the existing gaze estimation reviews and surveys are described in Section 2. A detailed review of the published deep learning-based gaze direction regression methods is presented in Section 3 and followed by a list of publicly available datasets for the corresponding task in Section 4. Furthermore, Section 5 showcases the results of the re-evaluated methods under both static and temporal test conditions. Finally, Section 6 provides the concluding remarks of the paper.

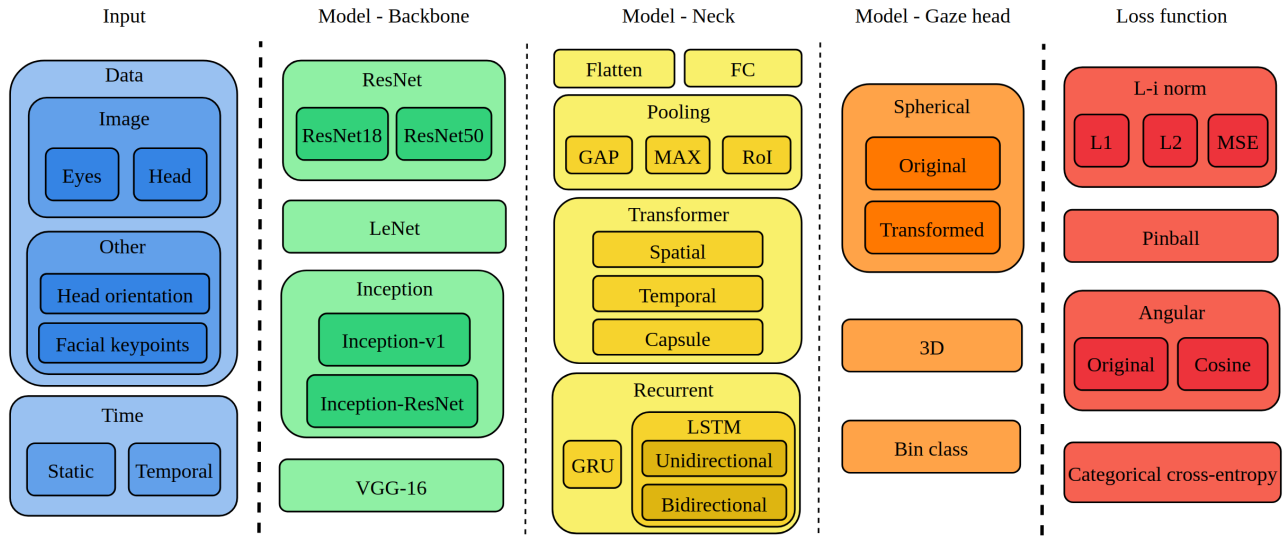

Figure 1. Component scheme of deep learning-based gaze regression solutions.

## 2. RELATED WORK

Pathirana et al. [3] published a survey on deep learning-based single- and multi-user approaches for gaze estimation. The survey analyzed architectures, datasets, coordinate systems, evaluation metrics, and environmental constraints. The authors designed criteria for the selection of an appropriate gaze estimation method when addressing a set of particular issues, and proposed a list of future research directions. Gosh et al. [4] presented a survey on automatic gaze analysis using deep learning-based methods. The paper presented a list of gaze estimation methods with respect to registration, representation, level of supervision, prediction, validation datasets, platforms, etc. Cheng et al. [5] published a review and benchmark for appearance-based gaze estimation using deep learning. Numerous methods were described in detail, divided by the features used (eye/facial images, videos), model types (supervised, self-/semi-/un-supervised, multi-task, recurrent, etc.), calibration methods (domain adaptation, user-unaware data collection), cameras used (single, multi, infrared, RGBD, etc.), and platforms used (computer, mobile phone, head-mounted device). In addition to presenting the results of various gaze direction and PoG estimation methods on different datasets, the benchmark also included a data rectification method, whose goal was to eliminate environmental factors and simplify the estimation task. Sharma and Chakraborty [6] presented a review of driver gaze estimation and its applications in gaze behavior understanding. Existing driver gaze benchmarks were listed alongside data collection methodologies and equipment. Gaze behavior understanding was focused on high-risk driving actions such as maneuvering through intersections, lane switching, etc. To the best of our knowledge, this is the first survey that focuses solely on deep learning-based methods for gaze direction regression.

## 3. DEEP LEANING-BASED GAZE DIRECTION REGRESSION

Figure 1 displays the general component scheme of deep learning-based solutions for gaze direction estimation from the literature. The scheme is designed with respect to the input data, model architecture (consisting of a backbone, neck, and gaze prediction head), and the loss function used to supervise the model during training. Additionally, Table 1 shows the particular components used in each of the analyzed methods.

### 3.1 Input data

Input data fed to gaze direction estimation models can generally be divided by data type and time. The main data type used in deep learning-based gaze direction estimation methods is an image. In one of the earliest methods, Zhang et al. [2] attempted to predict the gaze direction vector using eye images. However, most of the methods [7-18] published afterward relied on using head images, with some [7,17] using both head and eye images as input. In addition to images, other data can be put as input, such as the head orientation vector in [2] or facial landmarks in [7].

Table 1. Chronological overview of deep learning-based gaze direction estimation models. A question mark indicates missing information that could not be found in papers or official implementations. S, T, and Trans indicate static, temporal, and transformer, respectively.

| Model | Input | | Model architecture | | | Loss function |
|---|---|---|---|---|---|---|
| | Time | Data | Backbone | Neck | Gaze head | |
| Zhang [2] | S | Eye+head orientation | LeNet | Flatten+FC | Spherical | L2 |
| Palmero [7] | T | Head+eyes+ facial keypoints | VGG-16 | Flatten+FC+GRU | Spherical | L2 |
| Pinball Static [8] | S | Head | ResNet18 | GAP+FC | Spherical | Pinball |
| Pinball LSTM [8] | T | Head | ResNet18 | GAP+FC+BiLSTM | Spherical | Pinball |
| Zhang [9] | S | Head | ResNet50 | - | Spherical | L1 |
| MSA [10] | S | Head | ResNet18 | Max-pool+GAP+FC | Transformer spherical | Pinball |
| MSA+Seq [10] | T | Head | ResNet18 | Max-pool+GAP+FC | Transformed spherical | Pinball |
| Oh [11] | S | Head | Inception-v1 | Trans. (spatial)+GAP | ? | ? |
| GazeTR-Hybrid [12] | S | Head | ResNet18 | Trans. (spatial) | Spherical | L1 |
| GazeNAS-ETH [13] | S | Head | Custom (NAS) | Trans. (spatial)+GAP | ? | L1 |
| GazeCaps [14] | S | Head | ResNet18 | Trans. (capsules) | Spherical | MSE |
| L2CS-Net [15] | S | Head | ResNet50 | - | Bin class | CCE+MSE |
| MCGaze [16] | T | Head | ResNet50 | RoI pool+Trans. (spatial+temporal)+FC | 3D | Angular |
| CrossGaze [17] | S | Head+eyes | Inception-ResNet (head) +ResNet18 (eyes) | Trans. (spatial)+GAP | 3D | Cosine |
| Hybrid-SAM+LSTM [18] | T | Head | ResNet18 | Trans. (spatial)+LSTM | Spherical | Angular |
| Hybrid-SAM+Tx [18] | T | Head | ResNet18 | Trans. (spatial+temporal) | Spherical | Angular |

Regarding the time component, the input data can be given as static by using an image from a single timestamp or as temporal by using a sequence of consecutive image frames. In the temporal methods, the sequence length was set to: 4 in [7], 7 in [8,10,16], and 30 in [18].

### 3.2 Model architecture

Since each of the deep learning-based gaze estimation methods uses images as input, the first part of the architecture always comprises a backbone network that extracts visual features from the images. The most commonly used backbone is the ResNet [19] model. Particularly, ResNet18 was employed in [8,10,12,14,17,18], whereas ResNet50 was chosen in [9,15,16]. Models from the Inception [20] family have also been used in multiple solutions, such as Inception-v1 [20] in [11] and Inception-ResNet [21] in [17]. Furthermore, the LeNet [22] model was utilized in [2], whereas the VGG-16 [23] model was used in [7]. Unlike the aforementioned methods that use popular backbone networks, Nagpure and Okuma [13] performed a neural architecture search (NAS) to find the best-performing multi-resolution architecture for gaze direction estimation.

Next, the visual features extracted by the backbone network are combined with other input features (if any) and passed through the neck of the model. The neck of the gaze direction estimation models can start with the flattening operation, as in [2,7], or with various pooling operators: global average pooling (GAP) was utilized in [8], a combination of GAP and max-pooling in [10], and region-of-interest (RoI) pooling was performed in [16]. The flattening and pooling layers were followed by fully-connected (FC) layers in [2,7,8,10]. Furthermore, transformer encoder blocks have often been used in necks of gaze direction estimation models in recent years. These transformer blocks can be divided into three categories according to the type of features that their tokens model: spatial, temporal, and capsule. Spatial token-based transformers were employed in [11-13,17,18], whereas both spatial and temporal tokens were computed in [16,18]. As an alternative approach, Wang et al. [14] proposed a capsule token-based transformer that combines the concept behind capsule networks [24] with a self-attention mechanism. In addition, different recurrent units can be utilized to model temporal dependencies in methods that use a sequence of images as input: gated recurrent unit (GRU) was used in [7], unidirectional (past-only dependencies) long short-term memory (LSTM) in [18], and bidirectional (past and future dependencies) LSTM (BiLSTM) in [8].

Finally, the gaze prediction head is used to regress the gaze direction vector. The gaze vector is usually predicted as a 2D spherical vector consisting of yaw and pitch angles, as in [2,7-9,12,14,18]. Furthermore, in [10], the spherical gaze vectors were transformed using sine and cosine trigonometric functions. Alternatively, in [16,17], gaze was estimated as a 3D vector in the (*x*, *y*, *z*) form. Abdelrahman et al. [15] proposed using a classification head to simultaneously predict the bin class of gaze direction and regress the spherical gaze vector. Particularly, the yaw and pitch angles were classified into 4° bins, whereas the sum of bin probabilities obtained by the softmax activation function served as the prediction regression value.

### 3.3 Loss function

Various loss functions can be used to supervise the regression of the gaze vector. Many methods have focused on utilizing the L-i norm losses: L1 was used in [9,12,13], L2 was employed in [2,7], and mean squared error (MSE), the squared version of L2 loss, was used in [14]. In [15], the categorical cross-entropy (CCE) loss was coupled with the MSE loss. Furthermore, Kellnhofer et al. [8] proposed the Pinball loss function that simultaneously predicts the gaze vector and an offset value, which indicates the size of the cone around the ground truth within which the prediction should be with 80% confidence. The proposed loss was also used for supervision in [10]. In recent studies, angular-based losses, which calculate the angle between the predicted and ground truth gaze vectors, have gained increasing attention. The angular error (AE) loss is calculated by the following formula:

$$AE = arccos \frac{g \cdot \hat{g}}{\|g\| \|\hat{g}\|}, \tag{1}$$

where $g$ denotes the ground truth gaze vector and $\hat{g}$ denotes the predicted gaze vector. The loss function defined by (1) was used in [16,18], whereas the cosine loss function (the complement to 1 of $cos(AE)$) was utilized in [17].

## 4. DATASETS

Numerous gaze direction estimation datasets have been published over the last decade. Table 2 displays a list of publicly available datasets that can be utilized to train models for gaze direction regression with respect to the setting, number of subjects, data type, resolution, and total size. Most of the datasets contained up to 60 subjects, whereas only Gaze360 [8] and ETH-XGaze [9] had 100+ subjects. Early datasets were acquired in controlled (laboratory) settings, whereas newer datasets were collected in uncontrolled (in-the-wild) settings. In addition, there has been a noticeable shift from static (image) to temporal (video) datasets. The resolution of the images is generally full HD (1920x1080) or higher. Unlike other datasets, the EYEDIAP [26] and EVE [29] datasets were collected using two different resolutions. In this research, we decided to use the Gaze360 dataset, as it is an in-the-wild dataset with the largest diversity in the number of subjects and gaze vector distribution. The dataset is split into subject-wise stratified train, validation, and test *.txt* files, each of which contains a set of paths to the head frames and corresponding ground truth gaze vectors.

Table 2. Chronological overview of publicly available datasets for gaze direction regression.

| Dataset | Subjects | Setting | Data type | Resolution | Total frames |
|---|---|---|---|---|---|
| ColumbiaGaze [25] | 56 | Laboratory | Image | 5184x3456 | 5 880 |
| EYEDIAP [26] | 16 | Laboratory | Video | 640x480, 1920x1080 | 94x (2-3min) videos |
| UTMultiview [27] | 50 | Laboratory | Image | 1280x1024 | 64 000 |
| MPIIGaze [2] | 15 | Wild | Image | 1280x720 | 213 659 |
| RT-GENE [28] | 15 | Laboratory | Image | 1920x1080 | 122 531 |
| Gaze360 [8] | 238 | Wild | Video | 4096x3382 | 172 000 |
| EVE [29] | 54 | Laboratory | Video | 1920x1080, 1920x1200 | 12 308 334 |
| ETH-XGaze [9] | 110 | Laboratory | Image | 6000x4000 | 1 083 492 |
| GAFA [30] | 8 | Wild | Video | 1224x1024 | 882 000 |
| GFIE [31] | 61 | Wild | Video | 1920x1080 | 71 799 |

## 5. REPRODUCIBILITY AND DISCUSSION

In this research, we utilized each of the models from Table 1 that have the corresponding publicly available code. In the literature, models developed using the Gaze360 dataset were usually tested on some of the following subsets: the entire dataset (Full), gazes with a yaw angle up to $\pm 90°$ (Front), gazes with a yaw angle up to $\pm 20°$ (Front facing), gazes with a yaw angle from $\pm[90°-180°]$ range (Backward), and a subset of images in which a face can be detected (Detectable face). The existence of five different test subsets resulted in papers reporting results on one test subset and comparing them with the results of another method on a different test subset, making the comparison completely invalid. Some examples include the comparison of the Front results obtained by the L2CS-Net model in [15] with results of other methods that were in fact reported on the Detectable face subset, and the comparison of the Full results in [18], where the results of the Hybrid-SAM models were in fact obtained on the Detectable face subset. To address this issue, we trained the models using the entire dataset and evaluated them using each of the aforementioned test subsets.

Furthermore, the temporal results reported in some of the methods are not comparable with those of other studies, owing to changes in the evaluation protocol. Particularly, in the MCGaze [16] method, the result 7-frame sequences were stacked with a stride of 4, i.e., the last half of the previous sequence and the first half of the next sequence were averaged through the video, which effectively enlarged the temporal receptive field of the model. In the Hybrid-SAM [18] method, the evaluation metric was calculated as the mean of all frames in the sequence instead of the value of a single (usually central) frame, as in previous temporal methods [8,10]. Additionally, the aforementioned method used 30-frame sequences with frame $F$ from Gaze360 *.txt* files as the last frame instead of the 7-frame sequences with $F$ as the central frame, as in previous methods [8,10]. To address the problem of inconsistent temporal evaluation, we decided to use the same evaluation protocol for each of the tested temporal methods. In particular, we decided to follow the temporal evaluation protocol proposed in the paper that presented the Gaze360 dataset, which considered only the result of the central prediction frame from a 7-frame sequence and did not include any stacking. Following this decision, stacking was discarded in the MCGaze method. Also, for Hybrid-SAM models, changes included training and testing using 7-frame sequences (with $F$ as central) and the evaluation metric being calculated for $F$ only. Moreover, since the proposed Hybrid-SAM+LSTM model predicted gaze using 30 past frames and the adjusted model attempts to predict the central frame of 7 images, the BiLSTM-based model was tested alongside the proposed LSTM-based model.

Finally, in the L2CS-Net [17] method, an image size of 448x448 was used instead of a commonly used 224x224, thus giving the model the possibility to regress the gaze from four times larger images than in other methods. On the other hand, in the Hybrid-SAM method, a low resolution of 128x128 was utilized. To provide a fair comparison between methods, we decided to use a 224x224 image size for each model, including the L2CS-Net and Hybrid-SAM models.

Table 3. Results of static models (top), temporal models (middle), and temporal models under static test conditions (bottom). The best results from the static and temporal groups are underlined and bolded, respectively. Lower is better.

| Model | Full | Front | Front facing | Backward | Detectable face |
|---|---|---|---|---|---|
| Pinball Static [8] | 15.95 | 13.09 | 12.97 | 26.24 | 12.43 |
| MSA [10] | <u>13.90</u> | <u>12.23</u> | <u>12.25</u> | <u>19.90</u> | <u>11.46</u> |
| GazeTR-Hybrid [12] | 15.29 | 12.88 | 13.06 | 23.94 | 12.04 |
| L2CS-Net [15] | 15.81 | 13.12 | 13.14 | 25.49 | 12.38 |
| Pinball LSTM [8] | 13.68 | 11.44 | 11.32 | 21.75 | 10.72 |
| MSA+Seq [10] | **12.48** | **10.68** | **10.15** | **18.97** | **10.02** |
| MCGaze [16] | 13.13 | 11.14 | 10.79 | 20.29 | 10.70 |
| Hybrid-SAM+LSTM [18] | 17.02 | 13.28 | 12.38 | 30.50 | 12.10 |
| Hybrid-SAM+BiLSTM [18] | 17.10 | 13.28 | 12.56 | 30.83 | 12.09 |
| Hybrid-SAM+Tx [18] | 17.32 | 13.44 | 12.45 | 31.31 | 12.29 |
| Pinball LSTM [8] (Static cond.) | 15.01 | 12.73 | 12.68 | 23.22 | 11.87 |
| MSA+Seq [10] (Static cond.) | 13.60 | 11.86 | 11.36 | 19.88 | 10.91 |

The models were evaluated by using the mean AE across a particular testing subset. Table 3 displays the obtained results of the gaze direction regression models, with each result being the average of the five stochastic runs. The results of the static methods show that the MSA model outperformed all other methods by a significant margin, whereas the GazeTR-Hybrid model achieved the second-best result on most of the subsets. The results of the temporal methods show the superiority of the MSA+Seq model (temporal counterpart of the MSA model), whereas the MCGaze model obtained the second-best result on most of the subsets. Interestingly, the Hybrid-SAM+LSTM model achieved slightly better or similar results (depending on the subset) as the Hybrid-SAM+BiLSTM, although the prediction was performed for the central frame of a sequence instead of the last frame. In general, various temporally trained models, except Hybrid-SAM models, achieve better results than statically trained models, which implies the importance of temporal input on the gaze regression task.

Next, we analyze the performance of the temporal Pinball LSTM and MSA+Seq models, which have static counterparts (i.e., Pinball Static and MSA), under static test conditions for two main reasons. First, the acquisition setup of the Gaze360 dataset resulted in videos with continuous gaze movement. It is possible that temporal models achieve better results than static models because of overfitting to the continuum of gaze movement through the sequence. Such overfitting may lead to worse results of temporal models under static test conditions. Second, since human gaze is often static, focused on a particular interest for some time, performance analysis of temporal models under such conditions is of high importance. In this experiment, we passed 7 instances of the same image, which was previously placed in the middle of a 7-frame sequence, through models trained with temporal sequences. The results show that both models achieved better results than their static counterparts for each test subset.

## 6. CONCLUSION

Deep learning-based gaze direction regression is a diverse research field with numerous published methods. However, invalid comparisons are often made in these papers, mainly by comparing the results of different test subsets or by changing the evaluation protocol. Researchers should avoid making such mistakes for their methods to be a real contribution that can improve this field. The evaluation of numerous static and temporal methods using the same validation setup revealed that the MSA and MSA+Seq models, although more than 3 years old, significantly outperformed all the other methods on each of the testing subsets under static and temporal test conditions, respectively. Furthermore, the evaluation of temporal models under static test conditions showed the benefits of temporal training for generalization under static conditions.

## ACKNOWLEDGMENT

This work has been supported by Croatian Recovery and Resilience Fund - NextGenerationEU (grant C1.4 R5-I2.01.0001).

## REFERENCES

[1] E. Du Bois-Reymond. Untersuchungen über thierische elektricität: bd., 1. abth (Vol. 2, No. 1), G. Reimer, Berlin, (1849).

[2] X. Zhang, Y. Sugano, M. Fritz and A. Bulling, "Appearance-based gaze estimation in the wild," in Proc. IEEE Conf. Comput. Vis. Pattern Recognit., 4511-4520, (2015).

[3] P. Pathirana, S. Senarath, D. Meedeniya and S. Jayarathna, "Eye gaze estimation: A survey on deep learning-based approaches," Expert Syst. Appl. 199, 116894, (2022).

[4] S. Ghosh, A. Dhall, M. Hayat, J. Knibbe and Q. Ji, "Automatic gaze analysis: A survey of deep learning based approaches," IEEE Trans. Pattern Anal. Mach. Intell. 46(1), 61-84, (2023).

[5] Y. Cheng, H. Wang, Y. Bao and F. Lu, "Appearance-based gaze estimation with deep learning: A review and benchmark," IEEE Trans. Pattern Anal. Mach. Intell. (Early access), 1-20, (2024).

[6] P.K. Sharma and P. Chakraborty, "A review of driver gaze estimation and application in gaze behavior understanding," Eng. Appl. Artif. Intell. 133, 108117, (2024).

[7] C. Palmero, J. Selva, M.A. Bagheri and S. Escalera, "Recurrent cnn for 3d gaze estimation using appearance and shape cues," in Proc. Brit. Mach. Vis. Conf., 1-13, (2018).

[8] P. Kellnhofer, A. Recasens, S. Stent, W. Matusik and A. Torralba, "Gaze360: Physically unconstrained gaze estimation in the wild," in Proc. IEEE/CVF Int. Conf. Comput. Vis., 6912-6921, (2019).

[9] X. Zhang, S. Park, T. Beeler, D. Bradley, S. Tang and O. Hilliges, "Eth-xgaze: A large scale dataset for gaze estimation under extreme head pose and gaze variation," in Proc. Eur. Conf. Comput. Vis. Part V, 365-381, (2020).

[10] A. Ashesh, C.S. Chen and H.T. Lin, "360-degree gaze estimation in the wild using multiple zoom scales," in Proc. Brit. Mach. Vis., 1-14, (2021).

[11] J.O. Oh, H.J. Chang and S.I. Choi, "Self-attention with convolution and deconvolution for efficient eye gaze estimation from a full face image," in Proc. IEEE/CVF Conf. Comput. Vis. Pattern Recognit., 4992-5000, (2022).

[12] Y. Cheng and F. Lu, "Gaze estimation using transformer," in Proc. Int. Conf. Pattern Recognit., 3341-3347, (2022).

[13] V. Nagpure and K. Okuma, "Searching efficient neural architecture with multi-resolution fusion transformer for appearance-based gaze estimation," in Proc. IEEE/CVF Winter Conf. Appl. Comput. Vis., 890-899, (2023).

[14] H. Wang, J.O. Oh, H.J. Chang, J.H. Na, M. Tae, Z. Zhang and S.I. Choi, "Gazecaps: Gaze estimation with self-attention-routed capsules," in Proc. IEEE/CVF Conf. Comput. Vis. Pattern Recognit., 2668-2676, (2023).

[15] A.A. Abdelrahman, T. Hempel, A. Khalifa, A. Al-Hamadi and L. Dinges, "L2cs-net: Fine-grained gaze estimation in unconstrained environments," in Proc. Int. Conf. Frontiers Signal Proc., 98-102, 2023.

[16] Y. Guan, Z. Chen, W. Zeng, Z. Cao and Y. Xiao, "End-to-End Video Gaze Estimation via Capturing Head-Face-Eye Spatial-Temporal Interaction Context," IEEE Signal Proc. Lett. 30, 1687-1691, (2023).

[17] A. Cătrună, A. Cosma and E. Rădoi, "CrossGaze: A Strong Method for 3D Gaze Estimation in the Wild," arXiv preprint arXiv:2402.08316, (2024).

[18] S. Jindal, M. Yadav and R. Manduchi, "Spatio-Temporal Attention and Gaussian Processes for Personalized Video Gaze Estimation," arXiv preprint arXiv:2404.05215, (2024).

[19] K. He, X. Zhang, S. Ren and J. Sun, "Deep residual learning for image recognition," in Proc. IEEE Conf. Comput. Vis. Pattern Recognit., 770-778, (2016).

[20] C. Szegedy, W. Liu, Y. Jia, P. Sermanet, S. Reed, D. Anguelov, D. Erhan, V. Vanhoucke and A. Rabinovich, "Going deeper with convolutions," in Proc. IEEE Conf. Comput. Vis. Pattern Recognit., 1-9, (2015).

[21] C. Szegedy, S. Ioffe, V. Vanhoucke and A. Alemi, "Inception-v4, inception-resnet and the impact of residual connections on learning," in Proc. AAAI Conf. Artif. Intell. 31(1), 4278-4284, (2017).

[22] Y. LeCun, L. Bottou, Y. Bengio and P. Haffner, "Gradient-based learning applied to document recognition," Proc. IEEE 86(11), 2278-2324, (1998).

[23] K. Simonyan and A. Zisserman, "Very deep convolutional networks for large-scale image recognition," arXiv preprint arXiv:1409.1556, (2014).

[24] S. Sabour, N. Frosst and G.E. Hinton, "Dynamic routing between capsules," in Proc. Conf. Neur. Inform. Proc. Syst., 3856-3866, (2017).

[25] B.A. Smith, Q. Yin, S.K. Feiner and S.K. Nayar, "Gaze locking: passive eye contact detection for human-object interaction," in Proc. ACM Symp. User Interface Soft. Technol., 271-280, (2013).

[26] K.A. Funes Mora, F. Monay and J.M. Odobez, "Eyediap: A database for the development and evaluation of gaze estimation algorithms from rgb and rgb-d cameras," in Proc. Symp. Eye Track. Res. Appl., 255-258, (2014).

[27] Y. Sugano, Y. Matsushita and Y. Sato, "Learning-by-synthesis for appearance-based 3d gaze estimation," in Proc. IEEE Conf. Comput. Vis. Pattern Recognit., 1821-1828, (2014).

[28] T. Fischer, H.J. Chang and Y. Demiris, "Rt-gene: Real-time eye gaze estimation in natural environments," in Proc. Eur. Conf. Comput. Vis., 334-352, (2018).

[29] S. Park, E. Aksan, X. Zhang and O. Hilliges, "Towards end-to-end video-based eye-tracking," in Proc. Eur. Conf. Comput. Vis. Part XII, 747-763, (2020).

[30] S. Nonaka, S. Nobuhara and K. Nishino, "Dynamic 3d gaze from afar: Deep gaze estimation from temporal eye-head-body coordination," in Proc. IEEE/CVF Conf. Comput. Vis. Pattern Recognit., 2192-2201, (2022).

[31] Z. Hu, Y. Yang, X. Zhai, D. Yang, B. Zhou and J. Liu, "Gfie: A dataset and baseline for gaze-following from 2d to 3d in indoor environments," in Proc. IEEE/CVF Conf. Comput. Vis. Pattern Recognit., 8907-8916, (2023).